# Res-GCNN: A Lightweight Residual Graph Convolutional Neural Networks for Human Trajectory Forecasting

Yanwu Ge, Mingliang Song

{15201619084, songml17}@163.com

## Abstract

*Autonomous driving vehicles (ADVs) hold great hopes to solve traffic congestion problems and reduce the number of traffic accidents. Accurate trajectories prediction of other traffic agents around ADVs is of key importance to achieve safe and efficient driving. Pedestrians, particularly, are more challenging to forecast due to their complex social interactions and randomly moving patterns. We propose a Residual Graph Convolutional Neural Network (Res-GCNN), which models the interactive behaviors of pedestrians by using the adjacent matrix of the constructed graph for the current scene. Though the proposed Res-GCNN is quite lightweight with only about 6.4 kilo parameters which outperforms all other methods in terms of parameters size, our experimental results show an improvement over the state of art by 13.3% on the Final Displacement Error (FDE) which reaches 0.65 meter. As for the Average Displacement Error (ADE), we achieve a suboptimal result (the value is 0.37 meter), which is also very competitive. The Res-GCNN is evaluated in the platform with an NVIDIA GeForce RTX1080Ti GPU, and its mean inference time of the whole dataset is only about 2.2 microseconds. Compared with other methods, the proposed method shows strong potential for onboard application accounting for forecasting accuracy and time efficiency. The code will be made publicly available on GitHub.*

## 1. Introduction

As environmental pollution and traffic congestion increases, a lot of researches have been focused on ADVs, which is assumed to be the solution for those deteriorating problems. In order to achieve large-scale commercial application, the driving capability including efficiency and security of ADVs must be much better than that of human drivers. For quite a long time in the future, ADVs and human-driving vehicles will share the roads. Therefore, the ability to understand the driving scenes for ADVs is significantly important and necessary to attain safer and more efficient driving. The most important point of understanding the driving scenes is accurate predicted trajectories of other traffic agents around the ADV.

Compared with other techniques of ADVs such as perception, planning and control, the study of prediction technique started relatively later. However, it has been studied extensively in recent years and many promising methods are proposed. Our work focuses on trajectory forecasting of pedestrians, which is considered to be the most challenging task for three reasons. The first one is the social interaction [1], which means that when a pedestrian is moving in the environment, he/she interacts with other agents such as other pedestrians, cars, trees and so on a lot. For example, a pedestrian always bypasses static environmental obstacles like trees and dustbins, while yields to fast moving cars and a group of people. The second one is the multi-modal problem [2], which means that by avoiding collision when facing path conflict, a pedestrian has a lot of reasonable movements such as stop, moving to right or speeds up and pass. The last one is the randomness of the motion [3]. Compared with vehicles, pedestrians are less restricted by traffic regulations because they often walk in off-road area. Moreover, the walking direction and speed of a pedestrian constantly changes due to small inertia and flexible movements.

Previous studies of the problem are divided into three categories. The first one is non-deep learning methods, such as HMM (Hidden Markov Model) [4] and KF (Kalman Filter) [5]. These methods fail in modeling the social interaction between pedestrians, and gains a poor result due to the relatively simple models which has a weak non-linear modeling capability.

The second one is RNNs (Recurrent Neural Networks), such as LSTM (Long Short-Term Memory) networks [6, 7, 8] and GRU (Gated Recurrent Units) [10]. Commonly, these methods model every pedestrian by several RNNs and achieve social interaction by pooling or concatenation operation. However, RNNs are inefficient regarding of parameters size, which means that to achieve the same accuracy RNNs need more data to learn the pattern. Besides, the pooling or concatenation operation used to model social interaction is short of intuitive physical meaning.

The last one is GCNNs (Graph Convolutional Neural Networks) [9, 12, 13, 15, 18, 20]. These methods model the whole scene as unstructured data in the form of a graph, where the nodes in the graph represent entities and the edges denote the relative relationship of the connected nodes. GCNNs are parameter-efficient and time-saving for online inference. Most importantly, the graphs are capable of directly modeling the interactive patterns between pedestrians and taking impacts from all other pedestrians in the same scene into account. Therefore, we adopt GCNNs for forecasting trajectories of pedestrians in this work.

Specifically, we propose a novel robust trajectory forecasting network named Res-GCNN. We apply residual structure [14, 26] which consists of shortcut connections and convolution mapping. In this work, a trajectory is divided into two parts: the linear part and the nonlinear part. The linear motion involves no direction and speed change while the nonlinear motion changes the moving direction or speed. By employing residual structure, the linear motion is learned by the shortcuts which realizes identity mapping, while the nonlinear part is modeled by the convolution mapping. Meanwhile, a creative method is proposed to constructed the adjacent matrix for the graph of the studied scene. The novelties include speed and direction correction when calculating the kernel function between two interested pedestrians.

The rest of this paper is structured as follows: in Section 2, some closely related work is reviewed. The proposed Res-GCNN is introduced in Section 3 in detail. Experimental comparisons with the state-of-the-art trajectory prediction methods on benchmark datasets included ETH and UCY are presented in Section 4. Finally, the conclusion is drawn in Section 5.

## 2. Related work

A brief review of related work is given in this part. Considering the recent progresses, we focus on three aspects, which are described below.

### 2.1. RNNs for trajectory prediction

The Social-LSTM in [6] is the first research that takes the social interactions between proximal pedestrians into account. It applies LSTM to extract sequential position data pattern for each trajectory in a scene. Creatively, the pooling mechanism is proposed to pool the hidden states of the neighbors within a certain spatial distance. Then the pooled information is used as an input at the next time-step to fulfill temporal information share. However, only partial interactions by neighbor grids within a certain radius are calculated instead of global interactions.

The Social-GAN in [10] uses an encoder-decoder architecture. The encoder adopts a LSTM for each person and then the encoded information is connected to a novel pooling layer. The pooling layer computes the relative positions between the interested and all other people. These positions are concatenated with each person's hidden state, processed independently by an MLP (Multi-Layer Perceptron), then pooled elementwise to compute interested person's pooling vector. The predicted trajectory is generated based on the pooling vector.

The method in Meituan [11] proposes an interaction net which models the global interaction information. At the first phase, a LSTM is applied for each pedestrian to encode the trajectory feature. Then a max-pooling operation is used to model social interactions. The output of the max-pooling layer is connected to a GRU to further encode spatio-temporal feature of the current scene. When predicting future trajectories of some specific pedestrian, a query operation is designed to inquire about the social interaction.

### 2.2. Graph models for trajectory prediction

The STGAT in [22] adopts an encoder-decoder framework. LSTMs are used to extract the embedding, then the embedded features are connected to a graph attention model. For a specific scene the pedestrians are considered as nodes on the complete graph at every time-step, while the edges on the graph represent the exist of human-human interactions. Also, it allows a node to assign weighted importance to different nodes within a neighborhood and aggregate features from them.

The Social-BiGAT in [13] utilizes graph model to collect weighted attention of the neighbor pedestrians they interact with. A LSTM is connected behind the graph model to further extract deeper and wider properties.

The Social-STGCNN in [21] constructs the spatio-temporal graph for given frames of observation data. Then the convolution operation is performed on the constructed graph to create a spatio-temporal embedding. Following this, the time extrapolated convolutional neural networks predict future paths. Noticeably, the Social-STGCNN directly utilizes observation positions as graph nodes unlike its counterparts and several kernel functions are compared to calculate the graph edges.

Similar as in [21], we directly model the studied scenario by graph representation which is more intuitive and efficient than RNNs. Differently, we decouple a trajectory into linear and nonlinear features. Then we creatively adopt residual structure in our work to capture the linear feature from the input.

### 2.3. Social interactions modeling

As described above, there are two techniques to model social interactions commonly. One is the permutation invariant symmetric functions, such as max pooling or average pooling. The pooling method has two disadvantages, which are parameters inefficient and short of physical meaning. Apparently, the RNNs employed in the pooling method have more parameters than other network units, which further costs more time and data to train such models. On the other hand, the pooling method lacks intuitive sense for interactive behaviors.

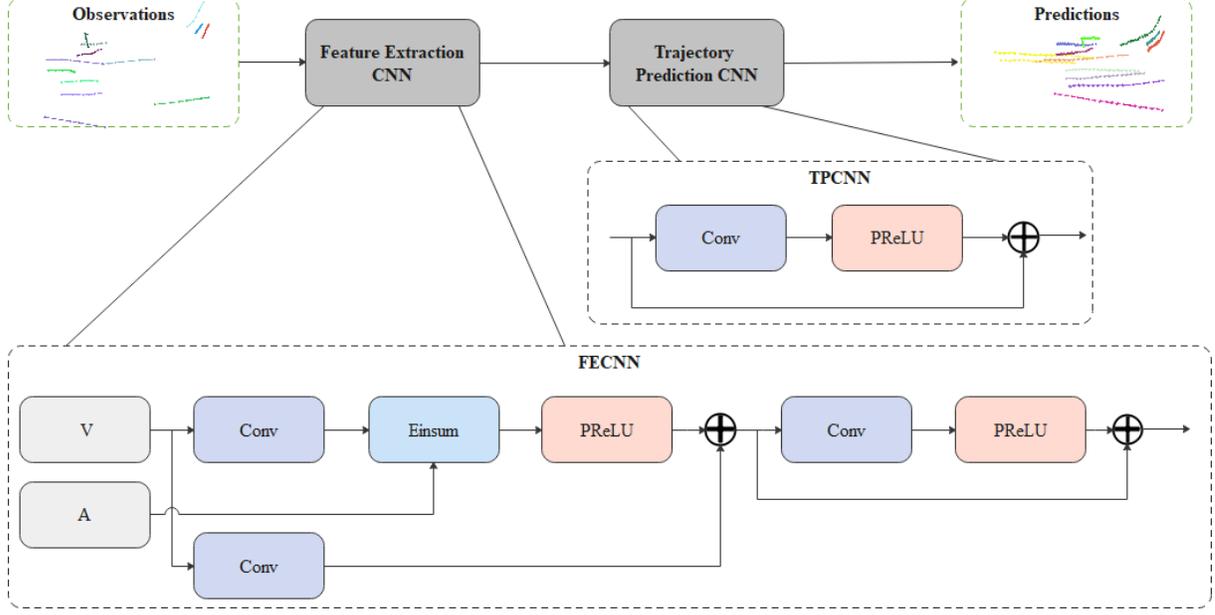

Figure 1: Overview of the proposed method. The proposed method contains two components, FECNN and TPCNN. FECNN plays the role of extracting spatio-temporal interaction feature, while TPCNN is designed to forecast the future trajectories. The Res-GCNN is composed of several FECNN layers and several TPCNN layers.

The other one is graph model which is mainly composed of nodes and edges. Typically, the features of pedestrians are employed as nodes and the relationship between pairs of features is employed as graph edge. The features could be history trajectories or extracted encoded information from history trajectories.

## 3. The proposed method

In this part we describe Res-GCNN in detail, including the model structure and the innovative speed and direction correction for the kernel function.

### 3.1. Problem Formulation

Suppose there are $N$ pedestrians walk in a typical scene like a university or a street. It needs at least $T_{obs}$ frames of observation data to predict the future data. For the $i$-th pedestrian at frame $t$, we denote its observation data as $Pos_i^t = \{x_i^t, y_i^t\}$, where $x$ and $y$ mean the longitudinal and lateral coordinates respectively, and $1 \leq i \leq N, 1 \leq t \leq T_{obs}$.

The purpose is to forecast the future trajectories for those pedestrians. Specifically, a trajectory consists of $T_{pred}$ path points, where each path point is composed of the longitudinal and lateral coordinates.

### 3.2. Feature Extraction Convolutional Neural Networks (FECNN)

The FECNN is designed to learn the intrinsic feature of the studied scene. The graph model is adopted for its capability to represent unstructured data. In the constructed graph, the vertices indicate the positions of the learned pedestrians, while the edges are designed to study the interactions of these pedestrians. We denote the graph model as:

$$G = \{V, A\} \quad (1)$$

where $V$ is the vertices tensor, while $A$ is the adjacent tensor.

At every time-step $t$, the vertices matrix $V^t$ is built as:

$$V^t = \begin{bmatrix} x_1^t & y_1^t \\ \vdots & \vdots \\ x_N^t & y_N^t \end{bmatrix} \quad (2)$$

Vertices tensor $V$ consists of all single frame vertices matrix $V^t$:

$$V = \{V^1, \cdots, V^{T_{obs}}\} \quad (3)$$

At every time-step $t$, the adjacent matrix $A^t$ is built as:

$$A^t = \begin{bmatrix} a_{11}^t & \cdots & a_{1N}^t \\ \vdots & \ddots & \vdots \\ a_{N1}^t & \cdots & a_{NN}^t \end{bmatrix} \quad (4)$$

where $a_{ij}^t$ denotes the interaction between the $i$-th pedestrian and the $j$-th pedestrian. The kernel function in [21] is upgraded to model $a_{ij}^t$:

$$a_{ij}^t = \begin{cases} \dfrac{1 * \omega_{dis} * \omega_{spd} * \omega_{dir}}{\sqrt{\left(x_i^t - x_j^t\right)^2 + \left(y_i^t - y_j^t\right)^2}}, i \neq j \\ 1, i == j \end{cases} \quad (5)$$

where $\omega_{dis}$ means the distance correction coefficient, $\omega_{spd}$ means the speed correction coefficient, $\omega_{dir}$ means the direction correction coefficient.

For $\omega_{dis}$, when pedestrian $j$ is out of the distance scope where pedestrian $i$ could reach in $T_{pred}$ frames, it is set to zero; otherwise, it is set to one:

$$\omega_{dis} = \begin{cases} 1, dis_{ij} \leq dis_{thres} \\ 0, dis_{ij} > dis_{thres} \end{cases} \quad (6)$$

where $dis_{thres}$ means the distance scope.

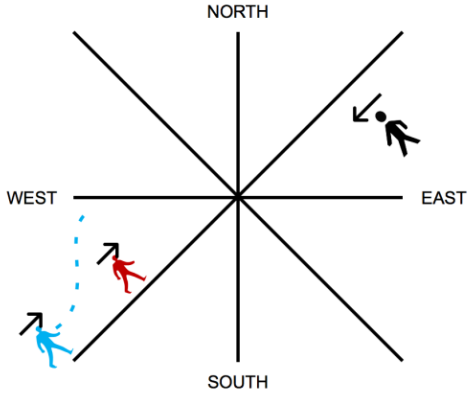

Figure 2. One representative social interacted scene with three pedestrians. The red one and the blue one moves towards one direction and the black one moves towards the opposite direction. Furthermore, the fast moving blue person is going to overtake the relatively slow moving red person.

The moving speed is divided into two levels, which are slow level and fast level, respectively. The speed threshold is denoted as $spd_{thres}$. For $\omega_{spd}$, when pedestrian $j$ (like the red one in Figure 2) moves in slow level in front of pedestrian $i$ (like the blue one in Figure 2) while pedestrian $i$ moves in fast level, it is set to a constant larger than one. Otherwise, it is set to one.

The moving direction is divided into eight partitions as shown in Figure 2. For $\omega_{dir}$, when pedestrian $j$ (like the red one in southwest) moves in the opposite direction of that of pedestrian $i$ (like the black one in northeast), it is set to a constant larger than one. Otherwise, it is set to one.

Then we symmetrically normalize each $A^t$ using the form in [12]:

$$A^t = (\Lambda^t)^{-\frac{1}{2}} (A^t + I)(\Lambda^t)^{-\frac{1}{2}}$$

where $\Lambda^t$ indicates the diagonal node degree matrix of $(A^t + I)$.

By described above, we complete the construction of the $V$ tensor and the $A$ tensor of a scene. Then we implement convolutional operation on the $V$ tensor to get the bi-variate Gaussian distribution [6] representation form which means that one dimension of the $V$ tensor is expanded from $\{x, y\}$ to $\{x, y, \mu_x, \mu_y, cov_{xy}\}$. After the transform, the **Einsum** operation is conducted to sum over weighted interactions from all other neighboring pedestrians at each frame. The **PReLU** function [27] is utilized as the activation function and we adopt the residual architecture. The above process is described as:

$$V_s = PReLU\big(Einsum(CNN(V), A)\big) + CNN(V) \quad (7)$$

where $V_s$ indicates the results of the spatial interactions.

Then we apply convolutional operation on $V_s$ to further extract temporal embedding. Also, the residual architecture is employed, which is:

$$V_{st} = PReLU\big(CNN(V_s)\big) + V_s \quad (8)$$

where $V_{st}$ indicates the results of the spatio-temporal interactions.

### 3.3. Trajectory Prediction Convolutional Neural Networks (TPCNN)

The TPCNN is designed to generate the future trajectory through the outcomes of FECNN. The outcomes include both the pedestrian motion feature and the social interactions with other traffic agents.

We directly adopt convolutional operation on the $V_{st}$ to get the forecasting trajectories with residual architecture again, which is:

$$V_{pred} = PReLU\big(CNN(V_{st})\big) + V_{st} \quad (9)$$

Noticeably, in the first TPCNN layer we need to expand one dimension in $V_{st}$ from $T_{obs}$ to $T_{pred}$ to generate the demanding frames of data.

| Model\Dataset | ETH | HOTEL | UNIV | ZARA1 | ZARA2 | AVG |
|---|---|---|---|---|---|---|
| Linear [6] | 1.33 / 2.94 | 0.39 / 0.72 | 0.82 / 1.59 | 0.62 / 1.21 | 0.77 / 1.48 | 0.79 / 1.59 |
| SR-LSTM-2 [8] | 0.63 / 1.25 | 0.37 / 0.74 | 0.51 / 1.10 | 0.41 / 0.90 | 0.32 / 0.70 | 0.45 / 0.94 |
| S-LSTM [6] | 1.09 / 2.35 | 0.79 / 1.76 | 0.67 / 1.40 | 0.47 / 1.00 | 0.56 / 1.17 | 0.72 / 1.54 |
| S-GAN-P [10] | 0.87 / 1.62 | 0.67 / 1.37 | 0.76 / 1.52 | 0.35 / 0.68 | 0.42 / 0.84 | 0.61 / 1.21 |
| SoPhie [17] | 0.70 / 1.43 | 0.76 / 1.67 | 0.54 / 1.24 | 0.30 / 0.63 | 0.38 / 0.78 | 0.54 / 1.15 |
| CGNS [25] | 0.62 / 1.40 | 0.70 / 0.93 | 0.48 / 1.22 | 0.32 / 0.59 | 0.35 / 0.71 | 0.49 / 0.97 |
| PIF [16] | 0.73 / 1.65 | **0.30** / 0.59 | 0.60 / 1.27 | 0.38 / 0.81 | 0.31 / 0.68 | 0.46 / 1.00 |
| STSGN [19] | 0.75 / 1.63 | 0.63 / 1.01 | 0.48 / 1.08 | 0.30 / 0.65 | 0.26 / 0.57 | 0.48 / 0.99 |
| GAT [13] | 0.68 / 1.29 | 0.68 / 1.40 | 0.57 / 1.29 | 0.29 / 0.60 | 0.37 / 0.75 | 0.52 / 1.07 |
| Social-BiGAT [13] | 0.69 / 1.29 | 0.49 / 1.01 | 0.55 / 1.32 | 0.30 / 0.62 | 0.36 / 0.75 | 0.48 / 1.00 |
| Social-STGCNN [21] | 0.64 / 1.11 | 0.49 / 0.85 | 0.44 / 0.79 | 0.34 / 0.53 | 0.30 / 0.48 | 0.44 / 0.75 |
| Meituan [11] | **0.39 / 0.79** | 0.51 / 1.05 | **0.25 / 0.56** | 0.30 / 0.61 | 0.36 / 0.73 | **0.36** / 0.75 |
| **Res-GCNN** | 0.65 / 1.12 | 0.31 / **0.50** | 0.38 / 0.72 | **0.28 / 0.50** | **0.24 / 0.41** | 0.37 / **0.65** |

Table 1. The ADE / FDE metrics of Res-GCNN compared with other counterparts. The unit is meter and the lower the better. We notice that Res-GCNN achieves the start-of-the-art results on multiple subsets.

## 4. Experimental results

Experimental comparisons with the state-of-the-art trajectory forecasting methods on datasets ETH and UCY are presented in this section. Besides, we perform an analysis about the inference results in detail.

### 4.1. Datasets Description

The proposed model is trained on the two well-known human trajectory prediction datasets, which are ETH [23] and UCY [24]. In ETH, two scenes named ETH and HOTEL are included. While in UCY, three scenes named ZARA1, ZARA2 and UNIV are contained. In these two datasets, the positions are observed every 0.4 second. The same as our superior counterparts, we use 8 historical data frames to forecast the coming 12 data frames. The evaluation metrics are ADE and FDE. The ADE is defined as the mean square error over all predicted trajectory points and the corresponding ground-truth trajectory points. Clearly, the ADE is used to measure the average prediction performance. The FDE is defined as the displacement error between the predicted location and the corresponding ground-truth location at the final frame named the $T_{pred}$ frame. The FDE is used to measure the location precision of the predicted terminal point. We use the leave-one-set-out strategy as in [6].

### 4.2. Model Parameters

Res-GCNN is composed of a series of FECNN layers followed by TPCNN layers. We set a training batch size of 128 and the model is trained for 200 epochs using Stochastic Gradient Descent (SGD). The initial learning rate is 0.01, and changed to 0.002 after 150 epochs. According to our ablation study on the numbers of network layers, the model which attains the optimal ADE and FDE metrics consists of 2 FECNN layers and 4 TPCNN layers.

### 4.3. Performance Evaluation

**ADE and FDE metrics** The experimental results are shown in Table 1, and we can see that Res-GCNN outperforms all the counterparts on the FDE metric by a large margin. Specifically, the FDE matric reduces from 0.75 meter to 0.65 meter which corresponds to a 13.3% improvement. As for the ADE metric, Res-GCNN achieves competitive results. Specifically, our results slightly deteriorate by 0.01 meter than state-of-the-art results by Meituan [11].

In detail, Res-GCNN gets the lowest prediction error on ZARA1 and ZARA2 subsets than all other methods. As for ETH and UNIV subsets, Res-GCNN presents competitive prediction accuracy. We want to emphasize the results on the HOTEL subset, which mainly consists of linear trajectories. As shown in Table 1, most counterparts get poorer generalization results than the pure linear method. Benefiting from the designed residual structure, our model attains the optimal FDE metric and the suboptimal ADE metric on this subset.

| Model\Metrics | Parameters size | Inference time |
|---|---|---|
| S-LSTM [6] | 264K | 1.179s |
| SR-LSTM-2 [8] | 64.9K | 0.158s |
| S-GAN-P [10] | 46.3K | 0.097s |
| PIF [16] | 360.3K | 0.115s |
| Social-STGCNN [21] | 7.6K | 0.002s |
| **Res-GCNN** | **6.4K** | **0.002s** |

Table 2. Experimental results on parameters size and inference time about some public counterparts compared to Res-GCNN. Models are bench-marked using Nvidia GTX1080Ti GPU.

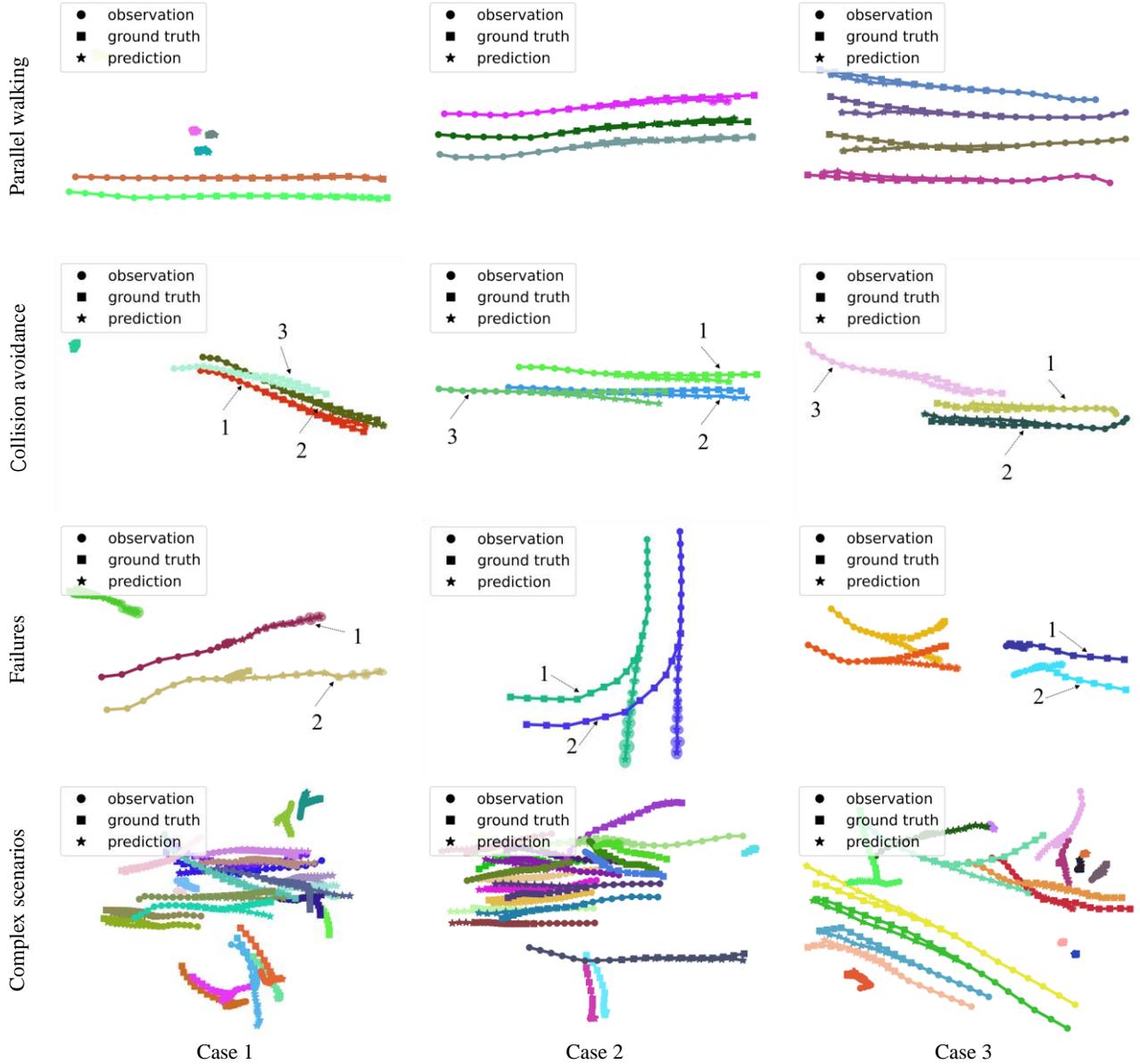

Figure 3. Visualization of forecasting results of four representative patterns. Each pattern consists of three scenes. The circle mark denotes eight frames of observation data. The square mark denotes 12 frames of ground truth data. The pentagram mark denotes 12 frames of forecasted trajectory points.

**Inference speed and model size** We aim to realize onboard application of Res-GCNN, therefore its parameters size and real-time performance are of great significance. Experimental results are shown in Table 2 [21]. For onboard application on ADVs, the consuming inference time should be less than 0.1 second per frame as the vehicles often move at a fast speed and need time to manipulate appropriate maneuvers to avoid collision. Fast inference speed reserves ADVs more time to perform such maneuvers. The parameters size determines the time cost of the model training process and small size means it consumes less time to complete training the model. From Table 1 we can see that our model has the least parameters and fastest inference speed, which gives Res-GCNN strong online ability.

### 4.4. Thorough analysis

The prediction results for some representative scenes are shown in Figure 3 and thoroughly analyzed in this subsec-tion. Some typical patterns including parallel walking, collision avoidance, complex interactions and failures are chosen. For each pattern, three cases are analyzed.

**Parallel walking** The first case comes from ZARA2 subset and there exist five pedestrians in the scene. Specifically, three of them are still and the other two walk abreast. The first figure in this pattern shows that the predicted trajectory perfectly matches the ground truth trajectory for both the still ones and the moving ones. The results prove that Res-GCNN successfully capture the interactions between these agents especially for the two parallel walking agents.

The second case comes from ZARA1 subset. There are three pedestrians in the scene and they move parallel with each other. The second figure in this pattern shows that the prediction results fit well with the ground truth, which means that Res-GCNN correctly models the social relations between them.

The last case also comes from the ZARA1 subset. There are four pedestrians in the scene and they are all in a parallel walking pattern. Similar as the results of the aforementioned two figures, the forecasting trajectory points coincide well with the ground truth ones, which means that Res-GCNN correctly models the mutual relations between them.

**Collision avoidance** All three cases in this pattern come from the ZARA1 subset. Three moving pedestrians exist in the first scene. Pedestrian 1 and 2 walk abreast and pedestrian 3 moves behind them. At the beginning pedestrian 3 locates on the right but at the end it locates on the left, named its path crosses with that of the other two. Our model achieves modeling the complex interactions during the whole process and generates collision-free trajectories which match the ground truth well.

In the second case, pedestrian 1 and 2 move parallel with each other and pedestrian 3 walk behind them. The ground truth shows that pedestrian 3 gradually changes its direction to the left until right behind pedestrian 2. It is noteworthy that the predicted trajectory gradually changes its direction to the right and avoids path overlapping. Though slightly deviates from the ground truth, the forecasting path is strongly plausible.

In the last case, pedestrians 1 and 2 walk parallel in one direction and the pedestrian 3 moves in the opposite direction. Res-GCNN derives satisfactory results which not only match the ground truth well but also achieves collision-free.

**Failure scenes** All three cases are selected from the ZARA1 subset in this pattern. In the first case we demonstrate a typical deceleration failure. The observed data shows that pedestrians 1 and 2 are moving at a nearly constant speed, but they suddenly perform a deceleration maneuver which results in poor forecasting performance.

In the second case we demonstrate a typical acceleration failure. The observed data shows that these two pedestrians are still, but they suddenly perform an acceleration maneuver which results in seriously deviated forecasting paths.

In the final case we demonstrate a typical direction change failure. In the historical sequences, the trajectories present no obvious intention for changing direction. However, in the coming sequences, it suddenly performs a turning, which results in seriously deviated forecasting trajectories.

**Complex scenarios** We choose all three cases in this pattern from the UNIV subset. As shown in the three pictures for this pattern, all cases consist of more than ten agents. Obviously, our model presents remarkable outcomes though these cases contain complex social interactions. The forecasting paths are mutual collision-free and coincide well with the real paths except for the unsuccessful patterns as previously mentioned.

## 5. Conclusion

An original method for pedestrian trajectory prediction is proposed in this manuscript. The novel method Res-GCNN is mainly composed of two submodules named FECNN and TPCNN. The FECNN submodule directly employees graph model to extract the scene feature, which consists of the trajectory information and the corresponding interactions between trajectories. The TPCNN submodule adopts convolutional operation on the spatio-temporal outcomes of the FECNN submodule to forecast the future trajectories. Both of them apply residual structure to resolve the learning of the linear patterns in the feature. Meanwhile, the learning of the nonlinear patterns is resolved by convolution mapping. It is noticeable that our model achieves the least parameter size and the fastest inference speed which give it strong real-time performance. Furthermore, Res-GCNN makes a significant improvement on the FED metric and attains a suboptimal ADE metric.

Future work will focus on extending Res-GCNN to other traffic agents such as cars, cyclists, trucks and so on. Moreover, more efforts will be paid to propose solutions to the three failure patterns.

## References

[1] Matthias Luber, Johannes A Stork, Gian Diego Tipaldi, and Kai O Arras. People tracking with human motion predictions from social forces. In 2010 *IEEE International Conference on Robotics and Automation*, pages 464–469. IEEE, 2010.
[2] Amirian, Javad , J. B. Hayet , and J. Pettre . "Social Ways: Learning Multi-Modal Distributions of Pedestrian Trajectories with GANs." (2019).
[3] Li, Xuesong , et al. "A recurrent attention and interaction model for pedestrian trajectory prediction." *IEEE/CAA Journal of Automatica Sinica* PP.99(2020):1-10.
[4] Qiao, S , et al. "A Self-Adaptive Parameter Selection Trajectory Prediction Approach via Hidden Markov Models." *IEEE Transactions on Intelligent Transportation Systems* 16.1(2015):284-296.
[5] Kim S , Guy S J , Liu W , et al. BRVO: Predicting pedestrian trajectories using velocity-space reasoning[J]. *International Journal of Robotics Research*, 2015, 34(2):201-217.

[6] Alexandre Alahi, Kratarth Goel, Vignesh Ramanathan, Alexandre Robicquet, Li Fei-Fei, and Silvio Savarese. Social lstm: Human trajectory prediction in crowded spaces. In *Proceedings of the IEEE conference on computer vision and pattern recognition*, pages 961–971, 2016.

[7] Huynh Manh and Gita Alaghband. Scene-lstm: A model for human trajectory prediction. *arXiv preprint arXiv:1808.04018*, 2018.

[8] Pu Zhang, Wanli Ouyang, Pengfei Zhang, Jianru Xue, and Nanning Zheng. Sr-lstm: State refinement for lstm towards pedestrian trajectory prediction. In Proceedings of the *IEEE Conference on Computer Vision and Pattern Recognition*, pages 12085–12094, 2019.

[9] Rianne van den Berg, Thomas N Kipf, and Max Welling. Graph convolutional matrix completion. *arXiv preprint arXiv:1706.02263*, 2017.

[10] Agrim Gupta, Justin Johnson, Li Fei-Fei, Silvio Savarese, and Alexandre Alahi. Social gan: Socially acceptable trajectories with generative adversarial networks. In Proceedings of the *IEEE Conference on Computer Vision and Pattern Recognition*, pages 2255–2264, 2018.

[11] Zhu Y, Ren D, Fan M, et al. Robust Trajectory Forecasting for Multiple Intelligent Agents in Dynamic Scene[J]. *arXiv*, 2020.

[12] Thomas N Kipf and Max Welling. Semi-supervised classification with graph convolutional networks. *arXiv preprint arXiv:1609.02907*, 2016.

[13] Vineet Kosaraju, Amir Sadeghian, Roberto Martń-Martń, Ian Reid, S Hamid Rezatofighi, and Silvio Savarese. Social-bigat: Multimodal trajectory forecasting using bicycle-gan and graph attention networks. arXiv preprint arXiv:1907.03395, 2019.

[14] K. He, X. Zhang, S. Ren, and J. Sun. Deep residual learning for image recognition. In 2016 *IEEE Conference on Computer Vision and Pattern Recognition (CVPR)*, pages 770–778, June 2016.

[15] Guohao Li, Matthias Muller, Ali Thabet, and Bernard Ghanem. Deepgcns: Can gcns go as deep as cnns? In Proceedings of the *IEEE International Conference on Computer Vision*, pages 9267–9276, 2019.

[16] Junwei Liang, Lu Jiang, Juan Carlos Niebles, Alexander G Hauptmann, and Li Fei-Fei. Peeking into the future: Predicting future person activities and locations in videos. In Proceedings of the *IEEE Conference on Computer Vision and Pattern Recognition*, pages 5725–5734, 2019.

[17] Amir Sadeghian, Vineet Kosaraju, Ali Sadeghian, Noriaki Hirose, Hamid Rezatofighi, and Silvio Savarese. Sophie: An attentive gan for predicting paths compliant to social and physical constraints. In Proceedings of the *IEEE Conference on Computer Vision and Pattern Recognition*, pages 1349–1358, 2019.

[18] Sijie Yan, Yuanjun Xiong, and Dahua Lin. Spatial temporal graph convolutional networks for skeleton-based action recognition. In Thirty-Second *AAAI Conference on Artificial Intelligence*, 2018.

[19] Lidan Zhang, Qi She, and Ping Guo. Stochastic trajectory prediction with social graph network. *arXiv preprint arXiv:1907.10233*, 2019.

[20] Boris Ivanovic and Marco Pavone. The trajectron: Probabilistic multi-agent trajectory modeling with dynamic spatio-temporal graphs. In The *IEEE International Conference on Computer Vision (ICCV)*, October 2019.

[21] Abduallah Mohamed, Kun Qian, Mohamed Elhoseiny, Christian Claudel. Social-STGCNN: A Social Spatio-Temporal Graph Convolutional Neural Network for Human Trajectory Prediction. The *IEEE/CVF Conference on Computer Vision and Pattern Recognition (CVPR)*, 2020, pp. 14424-14432.

[22] Huang, Yingfan, et al. "STGAT: Modeling Spatial-Temporal Interactions for Human Trajectory Prediction." 2019 *International Conference in Computer Vision* 2019.

[23] Stefano Pellegrini, Andreas Ess, Konrad Schindler, and Luc Van Gool. You'll never walk alone: Modeling social behavior for multi-target tracking. In 2009 IEEE 12th *International Conference on Computer Vision*, pages 261–268. IEEE, 2009.

[24] Alon Lerner, Yiorgos Chrysanthou, and Dani Lischinski. Crowds by example. In *Computer graphics forum*, volume 26, pages 655–664. Wiley Online Library, 2007.

[25] Jiachen Li, Hengbo Ma, and Masayoshi Tomizuka. Conditional generative neural system for probabilistic trajectory prediction. *arXiv preprint arXiv:1905.01631*, 2019.

[26] Orhan A E, Pitkow X. Skip Connections Eliminate Singularities[J]. *International conference on learning representations*, 2018.

[27] Kaiming He, Xiangyu Zhang, Shaoqing Ren, and Jian Sun. Delving deep into rectifiers: Surpassing human-level performance on imagenet classification. In Proceedings of the *IEEE international conference on computer vision*, pages 1026–1034, 2015.